\documentclass[letterpaper, 10 pt, conference]{ieeeconf}  

\IEEEoverridecommandlockouts                              

\usepackage{graphicx}
\usepackage{color,url}
\usepackage{hyperref}
\usepackage[normalem]{ulem}
\let\labelindent\relax
\usepackage{enumitem}
\usepackage{gensymb}
\usepackage{balance}
\usepackage{epsfig} 
\usepackage{amsmath} 
\usepackage{amssymb}  
\usepackage{url}
\usepackage{booktabs} 
\usepackage{cite}
\usepackage{subcaption}
\usepackage{caption}
\usepackage{graphicx}
\usepackage{xcolor}         
\usepackage{comment}
\usepackage{algorithm}
\usepackage{multirow}
\usepackage{array,booktabs}
\usepackage{algpseudocode}
\usepackage{dsfont}

\title{\LARGE \bf A Framework for Few-Shot Policy Transfer through Observation Mapping and Behavior Cloning
}

\author{Yash Shukla$^{1}$ Bharat Kesari$^{1}$ Shivam Goel$^{1}$ Robert Wright$^{2}$ Jivko Sinapov$^{1}$ %
\thanks{$^{1}$Yash Shukla, Bharat Kesari, Shivam Goel and Jivko Sinapov are with Department of Computer Science, Tufts University, Medford, Massachusetts, USA
{\tt\small $\{$yash.shukla; bharat.kesari; shivam.goel; jivko.sinapov$\}$@tufts.edu}}%
\thanks{$^{2}$Robert Wright is with the Georgia Tech Research Institute, Atlanta, Georgia, USA
         {\tt\small robert.wright@gtri.gatech.edu}}%
}

\begin{document}

\maketitle
\thispagestyle{empty}
\pagestyle{empty}

\newcommand{\rulesep}{\unskip\ \vrule\ }

\begin{abstract}
Despite recent progress in Reinforcement Learning for robotics applications, many tasks remain prohibitively difficult to solve because of the expensive interaction cost. Transfer learning helps reduce the training time in the target domain by transferring knowledge learned in a source domain. Sim2Real transfer helps transfer knowledge from a simulated robotic domain to a physical target domain. Knowledge transfer reduces the time required to train a task in the physical world, where the cost of interactions is high. However, most existing approaches assume exact correspondence in the task structure and the physical properties of the two domains. This work proposes a framework for Few-Shot Policy Transfer between two domains through Observation Mapping and Behavior Cloning. We use Generative Adversarial Networks (GANs) along with a cycle-consistency loss to map the observations between the source and target domains and later use this learned mapping to clone the successful source task behavior policy to the target domain. We observe successful behavior policy transfer with limited target task interactions and in cases where the source and target task are semantically dissimilar.
\end{abstract}


\section{Introduction}

Recent advances in Reinforcement Learning (RL) have enabled agents to learn the optimal behavior for a wide range of tasks, ranging from Atari games to autonomous driving~\cite{gulcehre2020rl,perez2022deep}. Despite these advancements, RL still suffers from sample-complexity issues, as the transition dynamics of the environment are unknown to the agent, and the agent needs to explore the environment to figure out the optimal behavior. Partially observable environments, sparse reward settings, and the dimensionality curse induced by large state and action spaces make many RL tasks prohibitively expensive to learn. This is a major concern in robotics that involves a costly and labor-intensive setup. 

Transfer Learning (TL)~\cite{taylor2009transfer} reduces the number of interactions required in a target domain by transferring relevant knowledge from a source domain.
Imitation learning~\cite{hussein2017imitation} is a type of TL that tries to mimic the policy provided by an expert user, thereby learning a policy in scenarios where the reward function of the task is inaccessible. For robotic settings, where training a behavior policy in a simulation environment is inexpensive, Sim2Real Transfer~\cite{hofer2021sim2real} is a special case of TL that allows a model trained in a simulation to be deployed on a physical robot. Training in simulation is easier as it does not have the physical time constraint of real-world settings. However, it suffers from ``Reality Gap''~\cite{tobin2017domain}, where the simulation policy performs poorly on transfer. Domain randomization requires access to a simulator where color, texture and other simulator parameters can be varied. This makes the learned policy fail on out-of-distribution test scenario. Continual learning on incrementally realistic simulations helps mitigate the Sim2Real problem as the agent attempts to transfer relevant information~\cite{josifovski2020continual}. One key limitation of Sim2Real approaches is the need for a simulation whose Markov Decision Process (MDP) representation exactly matches the complex dynamics of the realistic scenario, and is not always feasible~\cite{paull2020assessing}.

\begin{figure*}[t]
    \centering
    \subfloat[Few-Shot Policy Transfer Training Phase \label{fig:overview_training}]
    {\includegraphics[width=0.65\textwidth, height=0.42\textwidth]{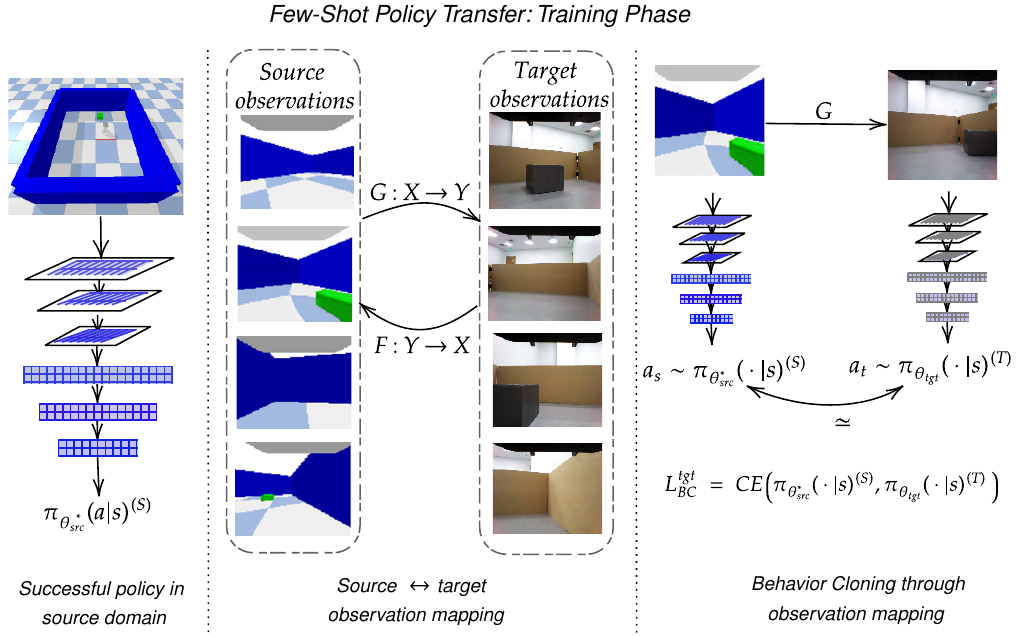}}
    \hspace{5mm}
    \rulesep
    \hspace{3mm}
    \subfloat[Evaluation Phase \label{fig:overview_evaluation}]
    {\includegraphics[width=0.15\textwidth, height=0.42\textwidth]{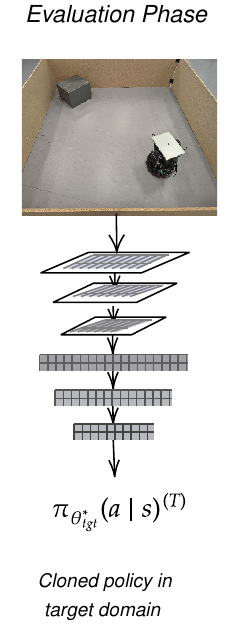}}    \caption{ Overview of the Few-Shot Policy Transfer approach. During training, the agent learns a mapping between the observations of the source and target domain and clones the successful source task behavior policy. During the evaluation, the agent tests the cloned policy on the target domain.}
    \label{fig:overview}
\end{figure*}

A major challenge in transferring knowledge from a source to a target task is finding a mapping between the two tasks~\cite{visus2021taxonomy}. Even with the access to a mapping, transferring knowledge remains difficult as the success of the transfer depends on the mapping, which may be inaccurate.
In this work, we propose a framework to achieve few-shot policy transfer from a simulated domain to a physical robotic domain. Our approach is task-agnostic as it does not require task-specific engineering to generate the mapping. It also does not require paired observations between the simulated and the physical domains to generate the mapping. Our approach to obtaining a successful behavior policy in the target task is shown in Figure~\ref{fig:overview}. We first obtain a successful behavior policy in the source domain using an off-the-shelf RL algorithm (e.g., PPO, DQN). Next, to obtain the observation mapping, we employ the CycleGAN~\cite{zhu2017unpaired} framework that enforces a pixel-level cycle-consistency loss to obtain the target domain observatioins given a simulated domain observation. Finally, using the source task policy and the observation mapping, we employ behavior cloning~\cite{ly2020learning} to adapt the successful source task policy to the target task. Recent works have addressed a task-agnostic framework for Sim2Real transfer~\cite{rao2020rl,ho2021retinagan}, and we extend this framework to a partially-observable setting that also contain semantic dissimilarities in the two domains.

Further, we compare how our approach scales when limited data is available from the target domain and when the target domain has different transition dynamics. This is necessary when the interactions in the target domain are costly and require an expensive setup. We conclude by scaling our approach to a target task that has different semantics and actions than the source task (e.g., the action 'move forward' moves the agent forward differently by different amounts).

\section{Related Work}
\textbf{Transfer Learning} (TL) uses knowledge from learned tasks and transfers it to a target task~\cite{taylor2009transfer,zhu2020transfer}. In RL settings, one popular TL technique is to transfer the source task value function parameters to initialize the value function of the target task~\cite{shukla2023agcl, konidaris2012transfer, lazaric2012transfer}. Policy transfer is another popular approach in which the policy learned in the source task is used to initialize the policy for the target task~\cite{kaspar2020sim2real,shukla2022acute, taylor2009transfer}.   

\textbf{Sim2Real Transfer} allows a model to be trained in simulation before deploying onto hardware, reducing time, cost, and safety issues. However, it encounters ``Reality Gap"~\cite{tobin2017domain}, where a model renders poor results on realistic domains due to changes in the environment parameters. The same authors introduce ``domain randomization'' as a solution, later expanded upon by Peng~\emph{et.~al.} \cite{peng2018sim}. In contrast, Operational Space Control \cite{kaspar2020sim2real} avoids domain randomization while speeding up training with fewer hyperparameters.
Domain adaptation strategy attempts at matching the representation in the two domains, allowing state knowledge to be transferred across semantically related tasks~\cite{you2019universal,xing2021domain}. Several works attempt to achieve domain adaptation by matching the image-based observations in the target domain to paired observations in the source domain~\cite{pan2017virtual,volpi2018adversarial,tl-da} using GANs~\cite{goodfellow2014generative} or unaligned GANs~\cite{zhu2017unpaired}. CycleGANs~\cite{zhu2017unpaired} are an extension of GANs that learn a mapping between two image domains without requiring paired data points in the two domains. This is beneficial for Sim2Real Transfer as obtaining paired observation matches is difficult when there are physical and semantic differences between the simulation and reality. Recent approaches have employed CycleGANs for robotic manipulation tasks~\cite{rao2020rl,ho2021retinagan} by jointly training an RL model along with the GAN or by introducing an object detection pipeline, improving manipulation accuracy. Unlike the aforementioned works, in this paper, we are interested in partially observable robotic navigation settings and in scenarios where the dynamics of the realistic domain can be semantically dissimilar to the dynamics of the simulation domain. We also intend to evaluate our approach on few-shot transfer, whereas the aforementioned approaches require re-training the learned policy in the real world.

\textbf{Behavior Cloning}~\cite{bain1999framework,ly2020learning} intends to learn a model that imitates expert demonstrations. In Behavior Cloning, the agent is provided with the raw observations and the actions taken by the expert demonstrator, and the algorithm trains a model that can predict the action given the observation during inference time. In BC, the agent does not need any additional training during test time and can generalize to novel in-distribution states during testing. Recent works have used BC for a wide range of applications, including training a quadcopter to fly along a path in a forest~\cite{rodriguez2019flying} and for autonomous driving~\cite{bojarski2016end,farag2018behavior}. Here, each observation is mapped with the corresponding action to perform behavior cloning. In the aforementioned works, the goal is to learn a policy for the same task as the expert demonstrations belong to. We are interested in cross-domain behavior cloning, where the task MDP of the target domain does not match the task MDP in which expert demonstrations are collected.

\section{Theoretical Framework}
In this section, we discuss the background information and how it fits into our few-shot policy transfer framework.
\subsection{Markov Decision Processes}
An episodic Markov Decision Process (MDP) $M$ is defined as a tuple $(\mathcal{S},\!\mathcal{A},\! p,\! r,\! \gamma, \mathcal{S}_0, \mathcal{S}_f)$, where $\mathcal{S}$ is the set of states, $\mathcal{A}$ is the set of actions, $p(s'|s,a)$ is the transition function, $r(s'\!,a,s)$ is the reward function and $\gamma\!\in\! [0,1]$ is the discount factor, and $\mathcal{S}_0$ and $\mathcal{S}_f$ are the sets of starting states and terminal states, respectively. At each timestep $t$, the agent observes a state $s$ and performs an action $a$ given by its policy function $\pi_\theta(a|s)$, with parameters $\theta$. The agent's goal is to learn an \emph{optimal policy $\pi_{\theta^*}\!$}, maximizing its discounted return $R_0 = \sum^{K}_{k = 0}\!\gamma^k\! r(s_{k+1}  ,a_k,s_k) $ until the end of the episode at timestep $K$. 

\subsection{CycleGAN}
Generative Adversarial Networks (GAN)~\cite{goodfellow2014generative} are a set of two networks, in which a generator and a discriminator contest with each other in a zero-sum game. The goal of the generator is to generate new fake datapoints that fool the discriminator, whereas the goal of the discriminator is to classify real datapoints from the fake ones produced by the generator. CycleGANs~\cite{zhu2017unpaired} are an extension of GANs that learn a mapping between two image domains $X$ and $Y$, through unpaired image samples $x \in X$ and $y \in Y$. In our experiments, the two observation spaces $X$ and $Y$ correspond to the observation datasets from source domain and the target domain, respectively. The role of the generators in CycleGAN is to learn two mappings, $G : X \rightarrow Y$, a mapping from domain $X$ to domain $Y$ and
$F : Y \rightarrow X$, a mapping from domain $Y$ to the domain $X$. While the role of the discriminators $D$ and $E$ is to distinguish real source task images $x$ from the adapted target task images $F(y)$; and the real target task images $y$ from the adapted source task images $G(x)$.
The loss for the source to target mapping is:
$$\mathcal{L}_{X\rightarrow Y}\!(\!G\!,\!E\!,\!X\!,\!Y\!)\!= \!\mathds{E}_{x \sim X}[log(1-E(G(x))] + \mathds{E}_{y \sim Y}[log E(y)]$$
Thus, the objective of the generator $G$ is to convert images from domain $X$ to resemble images from target domain $Y$ by minimizing the above objective, and the discriminator $E$ tries to maximize the above objective, giving us the optimization problem $min_G max_E \mathcal{L}_{X\rightarrow Y}(G,E,X,Y)$. The cycle-consistency loss of CycleGAN encourages pixel-wise consistency of images in the source task $x$ and the adapted source task image $F(G(x))$, and of the target task image $y$ and the adapted target task image $G(F(y))$:
\[\mathcal{L}_{cyclic}(G,F)= \!\mathds{E}_{x \sim X}d(x, F(G(x))) + \mathds{E}_{y \sim Y}d(y, G(F(y)))\]
where the distance metric $d$ is the mean-squared error. CycleGAN promotes generating fake datapoints that lie in distribution, without access to a paired mapping function.

\subsection{Behavior Cloning}
Behavior Cloning (BC) aims to learn a policy for a task that matches the given expert demonstrations. Using the expert demonstrations (set of $N$ state-action pairs $\{(s_i,a_i)_{i=1}^{N}\}$), BC performs maximum likelihood estimation, using a classifier or a regressor to determine the parameters of the model that would fit the expert demonstrations. 
The maximum likelihood problem of BC can be defined as:
\[\theta^*= \operatorname*{argmax}_{\theta} \prod_{i=0}^N \pi_{\theta}( a_i|s_i) \]
where $\theta$ are the parameters of the network $\pi_{\theta}$. The goal of the gradient in BC is to change $\theta$ such that the probability of the expert action ($a_i$) increases in the imitation policy’s distribution $\pi_{\theta}(\cdot|s_i)$, based on the expert's policy distribution. 
\section{Problem Formulation}


Let us consider a source task $M^{(S)}$ defined using the MDP $(\mathcal{S}^{(S)},\!\mathcal{A}^{(S)},\! p^{(S)},\! r^{(S)},\! \gamma^{(S)}, \mathcal{S}^{(S)}_{0}, \mathcal{S}^{(S)}_{f})$. Similarly, the target task $M^{(T)}$ is defined using the MDP $(\mathcal{S}^{(T)},\!\mathcal{A}^{(T)},\! p^{(T)},\! r^{(T)},\! \gamma^{(T)}, \mathcal{S}^{(T)}_{0}, \mathcal{S}^{(T)}_{f})$.

The source and the target domains $S$ and $T$ have different set of states $\mathcal{S}^{(S)} \neq \mathcal{S}^{(T)}$. The set of actions in the source and the target domains need not be consistent, but are defined using a mapping $\mathcal{X}_\mathcal{A}$ that maps actions in the source task $\mathcal{A}^{(S)}$ to actions in the target task $\mathcal{A}^{(T)}$. We assume similar transition dynamics and reward functions in the two domains; $p^{(S)} \approx p^{(T)}$ and $r^{(S)} \approx r^{(T)}$.

The agent also has access to two trajectory datasets, one in the source domain $\mathcal{D}^{(S)} = \{(s^{i}_t, a^{i}_t, s^{i}_{t+1})_{i = 1}^n\}$ and the other in the target domain $\mathcal{D}^{(T)} = \{(s^{i}_t, a^{i}_t, s^{i}_{t+1})_{i = 1}^m\}$, where $n$ and $m$ are the number of datapoints in the two datasets. The datasets are collected using randomly initialized behavior policies and may not encapsulate all possible states in the two domains. Additionally, the learning agent does not have access to the policies that collected the datasets. In this work, each state in the datasets is an egocentric image collected by a camera mounted on the robot. The trajectory dataset in the target domain is static, i.e. no further interactions are possible in the target domain for training a policy in the target domain. We make these assumptions to address real-world situations where the learning agent only has access to an offline dataset, and interacting with the target domain requires an expensive setup. 
The goal of the agent is to learn a beahvior policy in the source task and transfer it to the target task such that the zero-shot performance of the transferred policy achieves expected episodic return $R^{(T)} \geq \delta$ in the target task, where $\delta$ is predetermined threshold performance.



\section{Few-shot Policy Transfer Approach}\label{sec:approach}

The Few-Shot Policy Transfer Approach consists of three sections: (1) Obtaining a successful behavior policy for the source task; (2) Generating a mapping between the source and target task observations, and (3) Cloning the source task policy to the target task through the mapping.

The overall approach of our method is given in Algorithm~\ref{alg:zspt}. The goal is to transfer a behavior policy from a source task to a target task that has semantic and observational dissimilarities. The agent has access to the source task simulator and also has access to two static datasets (one in the source domain and one in the target domain) collected from a randomly initialized behavior policy. 

First, we use Proximal Policy Optimization~\cite{DBLP:journals/corr/SchulmanWDRK17} to learn a successful behavior policy for the task in the source domain (line 1). In the source domain, the cost of interactions is low (e.g., a simulator) and hence we can reasonably learn a behavior policy for the task. The RL algorithm takes in the egocentric image view of the robot as the input and returns the action that the robot should take according to the policy.
The next step entails learning an observation mapping between the states provided in the source and target domain datasets. Given the two static datasets $\mathcal{D}^{(S)}$ and $\mathcal{D}^{(T)}$, where each state in the dataset is an image, we use CycleGAN to learn the mapping between the images (line 2). CycleGAN does not require paired image data between the two domains to learn a mapping, and can be used when datasets are collected using randomly initialized policies. Additionally, the mapping does not assume a temporal relationship between the individual samples;  which allows us to use i.i.d datasets. The output of the learned mapping function is a tuple consisting of the two generators and two discriminators. The two generators, $G$ and $F$, provide a source$\rightarrow$target and a target$\rightarrow$source mapping respectively, whereas the two discriminators, $D$ and $E$ learn a classification between a real image captured by the camera on the robot and a fake one generated by the generators. Next, we incorporate this mapping and learn a target task policy by cloning the source task policy. Cloning the policy from the source task to the target task involves minimizing the action probability distribution mismatch between the learned source task policy and the target task policy. 

At each time step, the source task state from the simulator is passed through the mapping generator $G$ to produce a corresponding state in the target task (line 7). The ground truth action probability distribution for this generated target task state is the action probability distribution provided by the learned source task policy. Since the transition dynamics and the goal of the tasks in the source and the target domain are approximately equivalent, the resulting states of the agent in the source and the target domain after taking equivalent actions should have correspondence (line 10-12). Thus, the behavior cloning loss is simply the cross entropy loss between the source task and target task action probabilities. Through the mentioned few-shot policy-transfer approach, we intend to generate a target task policy whose output action distribution matches the source policy's action distribution. When the behavior cloning loss converges to a predetermined threshold value, we attain a successful target task behavior policy and then evaluate this policy on the target domain.\footnote{Our code base is available and can be found at: \href{https://github.com/shukla-yash/Few-Shot-Policy-Transfer.git}{https://github.com/shukla-yash/Few-Shot-Policy-Transfer.git}}. 
\begin{algorithm}[t]
\footnotesize
\caption{{\tt Policy\_Transfer}($M^{(S)}, \mathcal{D}^{(S)}, \mathcal{D}^{(T)}, \delta_{BC}$ )}
\label{alg:zspt}
\raggedright \textbf{Output}: 
Target task behavior policy: $\pi_{\theta^*_{tgt}}$\\
\textbf{Algorithm:}
\begin{algorithmic}[1] 
\State $\pi_{\theta^*_{src}} \leftarrow {\tt Learn}(M^{(S)})$\Comment{Learn a source task behavior policy}
\State $G, F, D, E \leftarrow {\tt CycleGAN}(\mathcal{D}^{(S)}, \mathcal{D}^{(T)})$\Comment{Learn a source $\leftrightarrow$ target observation mapping through CycleGAN}
\State $\pi_{\theta_{tgt}} \leftarrow {\tt Initialize\_Policy}()$
\While {True}
\State $s^{(S)} \leftarrow {\tt Environment.Reset}(M^{(S)})$
\While {True}
\State $s^{(T)} \leftarrow G(s^{(S)})$ \Comment{Adapted State in target domain}
\State $L^{tgt}_{BC}  =  {\tt Cross-Entropy(}\pi_{\theta^*_{src}}(\cdot|s^{(S)}),\pi_{\theta_{tgt}}(\cdot,s^{(T)})) $ 

\Comment{Minimizing action distribution mismatch}
\State ${\pi_{\theta_{tgt}} \leftarrow \tt Update\_Network}(L^{tgt}_{BC},\pi_{\theta_{tgt}})$
\State $s^{(S)}, rew, done = {\tt Environment.Step}(s^{(S)}, a^{(S)})$
\If {$done$}
\State ${\tt break}$
\EndIf
\EndWhile
\If {$L^{tgt}_{BC}<\delta_{BC}$}
\State $\pi_{\theta^*_{tgt}} \leftarrow \pi_{\theta_{tgt}}$
\State ${\tt break}$
\EndIf
\EndWhile
\State \textbf{return} $\pi_{\theta^*_{tgt}}$
\end{algorithmic}
\end{algorithm}

\section{Experiments and Results}

Through experiments, we aim to answer: (1) Can we perform a successful few-shot policy transfer when there exists a mismatch in the observations between two domains? (2) Given the costly interactions in the target domain, how does this approach scale to different numbers of interactions? (3) Can it scale to target environments that are semantically different in the observations and the action spaces? \\ (4) How does this approach perform for Sim2Real transfer?

\begin{figure}[t]
    \centering
    \subfloat[Source domain observations \label{fig:exp1-1}]
    {\includegraphics[width=0.235\textwidth, height=0.18\textwidth]{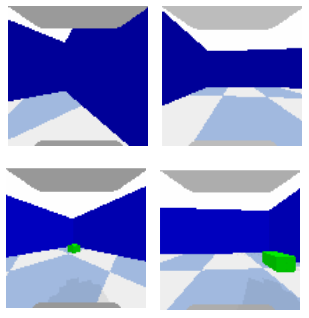}}
    \hspace{1mm}
    \subfloat[Target domain observations \label{fig:exp1-2}]
    {\includegraphics[width=0.235\textwidth, height=0.18\textwidth]{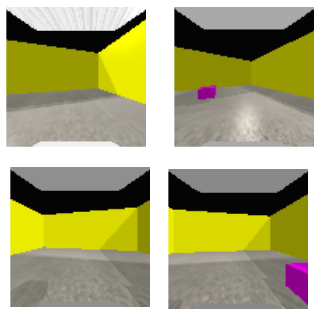}}    
    \caption{Examples of source and target domain observations.}
    \label{fig:exp1_obs}
\end{figure}

\subsection{Sim2Sim transfer with observation mismatch}\label{sec:exp_1}
To answer the first question, we evaluated our few-shot policy transfer approach on two simulated domains with a significant mismatch in the observations. In both domains, the goal of the agent is to reach the goal location, which is denoted by a box in the environment. The environment is partially observable, i.e., at each time step, the agent observes an egocentric view image using the camera mounted on the robot (Fig~\ref{fig:exp1_obs}). The agent is a Turtlebot acting in a simulated environment, implemented using PyBullet~\cite{coumans2019}. The set of discrete navigation actions available for the robot are moving forward and backward by $0.08m$, and turning left and right by $\frac{\pi}{10}$ radians. In each episode, the goal is spawned randomly within the bounds of the environment ($[2m\times2m]$). The agent acts in a sparse reward setting, and achieves a positive reward of $+1000$ once it reaches the goal object, and receives a $-1$ reward at all other timesteps. In this episodic setting, the episode ends when the robot achieves the goal, or reaches the maximum permissible timesteps (100) in an episode. For the first experiments, the only difference between the source and the target domains is a mismatch in the pixel-level agent observations, as seen in Fig~\ref{fig:exp1_obs}. RL agents are very fragile to changes in the observations. Even a minor pixel level change can lead to catastrophic results~\cite{qu2020minimalistic}. Thus, a direct policy transfer from the source to the target domain yields poor results.

We begin by learning a policy that achieves a $98\%$ success rate on the source task shown in Fig~\ref{fig:overview_training}. To learn the policy, we use Proximal Policy Optimization~\cite{DBLP:journals/corr/SchulmanWDRK17}. The structure of the model involves three convolutional layers of size $64$ followed by three linear layers\footnote{PPO Hyperparameters in Appendix~\href{https://github.com/shukla-yash/Few-Shot-Policy-Transfer/blob/main/appendix.pdf}{https://github.com/shukla-yash/Few-Shot-Policy-Transfer/blob/main/appendix.pdf}}. Once we have a successful source task policy, the next step entails learning a mapping between the source and target dataset observations using CycleGAN. The source and target dataset consisted of $4000$ state images taken using a random behavior policy on the two domains. CycleGAN outputs two generator functions that map images between the source and target domains. Employing this mapping function, we perform behavior cloning to learn the policy for the target domain, as defined in Section~\ref{sec:approach}. The experiments were conducted using a 64-bit Linux Machine, with Intel(R) Core(TM) i9-9940X 3.30GHz processor, 126GB RAM, and a NVidia RTX 2080 GPU.  
\begin{table}
    \centering
\begin{tabular}{c|c}
\small \textbf{Source-to-Target Transfer Approach}     &  \small \textbf{Success Rate}\\[0.3em]\midrule
\small Direct Policy Transfer     & \small $2\%$\\[0.3em]
\small GAN+Behavior Cloning & \small $68\%$\\[0.3em]
\small CycleGAN+Behavior Cloning (Ours) & \small $94\%$\\[0.3em]
\end{tabular}  
\caption{ Success rate comparison of different Source-to-Target Transfer Approaches. For the GAN+Behavior Cloning, and for CycleGAN+Behavior Cloning, 4000 source and target task images were used for generating the mapping. \label{fig:first_exp}}
\end{table}

The results from the experiments are shown in Table~\ref{fig:first_exp} . We observed that the few-shot performance of the transfer yielded a success rate of $94\%$ when evaluated on $100$ episodes in the target domain. We compared this approach with two baseline approaches, a direct policy transfer approach, where the trained source policy was directly transferred to the target task, and a few-shot policy transfer through GAN and behavior cloning. The results from the table show that a few-shot policy transfer through Cycle-GAN achieves better accuracy as compared to the baseline approaches. The cycle-conistency loss acts as a regularization for the generators, and guides the image generation process in the target domain closer toward image translation. The baseline GAN+BC did not employ the cycle-consistency loss while learning a mapping between the source and target images. This led to poor mapping and hence a poor policy in the target domain. Direct policy transfer did not show any reasonable success in the target domain.

\subsection{Comparison with different numbers of interactions}
\begin{figure}[t]
    \centering
    \includegraphics[width=0.35\textwidth, height=0.22\textwidth]{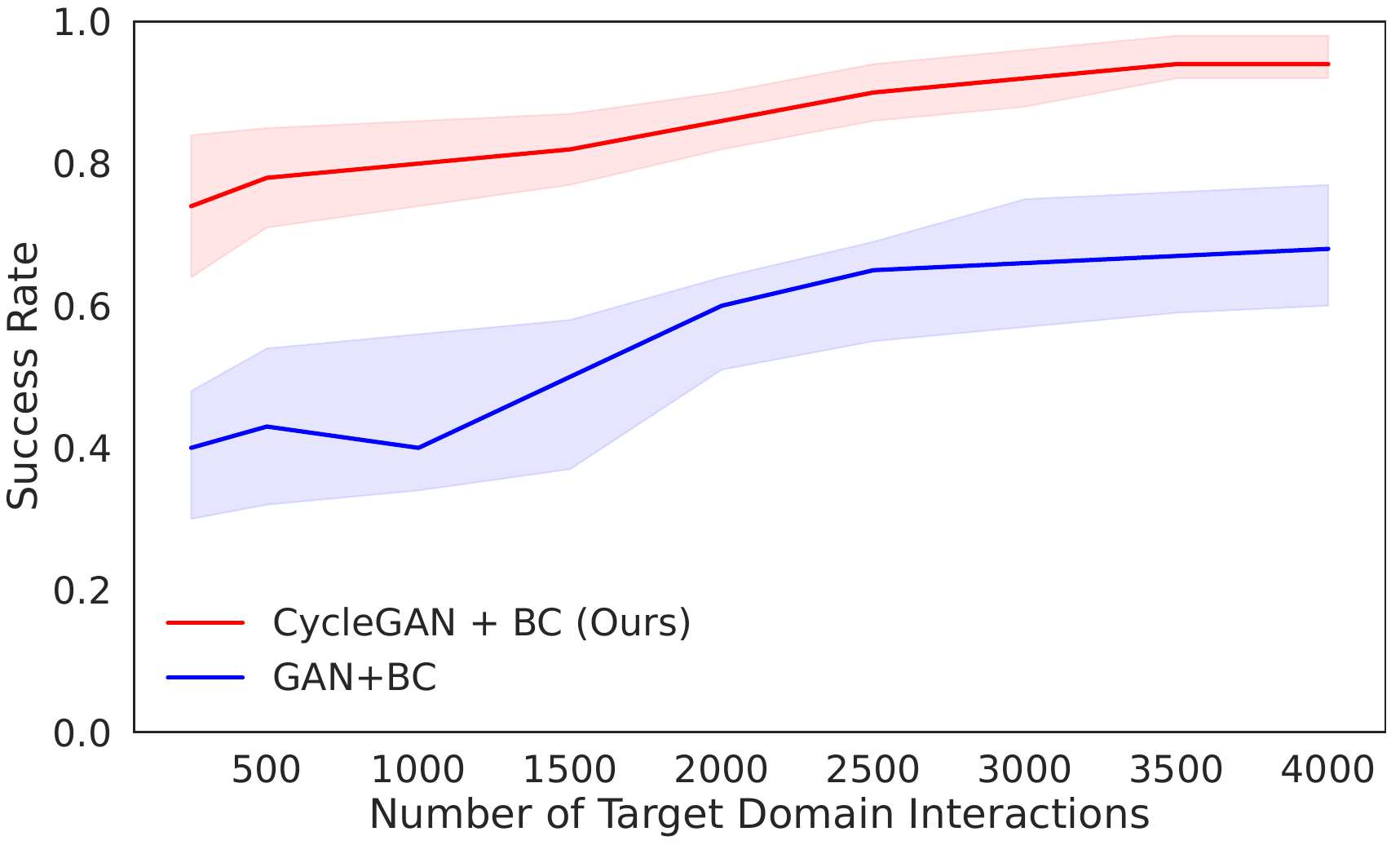}
    \caption{Success rate (Mean $\pm$ SD) comparison of the three TL approaches for different target dataset sizes.}
    \label{fig:second_exp}
\end{figure}

To answer the second question, we generate a mapping between the observations using different sizes of target datasets. The objective of this experiment is to determine how our proposed few-shot policy transfer approach compares when the number of datapoints available in the target dataset is limited. This is particularly useful when interactions in the target dataset require an expensive setup or are not easily attainable. For this experiment, we trained separate CycleGAN models with varying numbers of images in the target dataset, but a fixed number of images in the source dataset (4000 images).
The results of this experiment are shown in Fig~\ref{fig:second_exp}. As the size of the target dataset increases, the success rate of the task in the target domain increases, thus verifying that more images in the target dataset translate to a better mapping which in turn translates to an accurate target task representation. However, we see that even with only $250$ interactions, the few-shot policy transfer approach achieves a $74\%$ success rate, which can be improved with online fine-tuning on the target task. Our proposed approach outperformed the other baseline approach, GAN+BC, which did not employ the cycle-consistency loss. The number of datapoints required to achieve a successful mapping increases as the number of objects in the task increases. From our experiments, $\sim$200 images per object are sufficient for achieving a good mapping.


\begin{figure}[b]
    \centering
    \subfloat[Source observation \label{fig:exp3-1}]
    {\includegraphics[width=0.21\textwidth, height=0.16\textwidth]{src.png}}
    \hspace{3mm}
    \subfloat[Target observations \label{fig:exp3-2}]
    {\includegraphics[width=0.21\textwidth, height=0.16\textwidth]{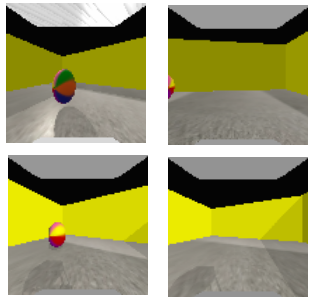}}    
    \caption{Examples of source and target domain observations.}
    \label{fig:exp3_obs}
\end{figure}

\subsection{Semantically different source and target environments}

To answer the third question, we evaluated our few-shot policy transfer approach on a target domain that is semantically different than the source domain in the observation and the action spaces. In this target domain, the goal of the agent is still to navigate to a goal object in the environment, but the object is a spherical ball that does not have a fixed base, as shown in Fig~\ref{fig:exp3_obs}. Whereas, in the source domain, the goal is to navigate to a box, with observations shown in Fig.~\ref{fig:exp1-1}. Unlike the goal object in the source task, the spherical ball produces different images when observed from different angles. This makes it difficult to attain a successful mapping with a limited number of images. Additionally, in the target task, the actions have finer discretizations, with a forward and backward movement causing the agent to move $0.03m$, and the rotation movements causing the agent to rotate $\frac{\pi}{15}$ radians instead of the actions described in section~\ref{sec:exp_1}. The source and target dataset consisted of $4000$ images taken using a random behavior policy on the two domains. The results from this experiment are summarized in Table~\ref{fig:third_exp}. We observe that the direct policy transfer is unable to generalize to the target task. The semantic difference between the two domains in terms of observations, actions, and transitions makes it very difficult for direct policy transfer and GAN+BC to accomplish the target task. Our proposed few-shot approach performs much better than the baseline approaches and achieves a $78\%$ success rate on the target task.
\begin{table}[ht]
    \centering
\begin{tabular}{c|c}
\small \textbf{Source-to-Target Transfer Approach}     &  \small \textbf{Success Rate}\\[0.3em]\midrule
\small Direct Policy Transfer     & \small $0\%$\\[0.3em]
\small GAN+Behavior Cloning & \small $20\%$\\[0.3em]
\small CycleGAN+Behavior Cloning (Ours) & \small $78\%$\\[0.3em]
\end{tabular}  
\caption{Success rate comparison of different Source-to-Target Transfer Approaches. For the GAN+Behavior Cloning, and for CycleGAN+Behavior Cloning, 4000 source and target task images were used for generating the mapping. \label{fig:third_exp}}
\end{table}
\subsection{Sim2Real Transfer on a Physical TurtleBot}
\begin{figure}[t]
    \centering
    \subfloat[Real-world domain \label{fig:real-world-domain}]
    {\includegraphics[width=0.23\textwidth, height=0.15\textwidth]{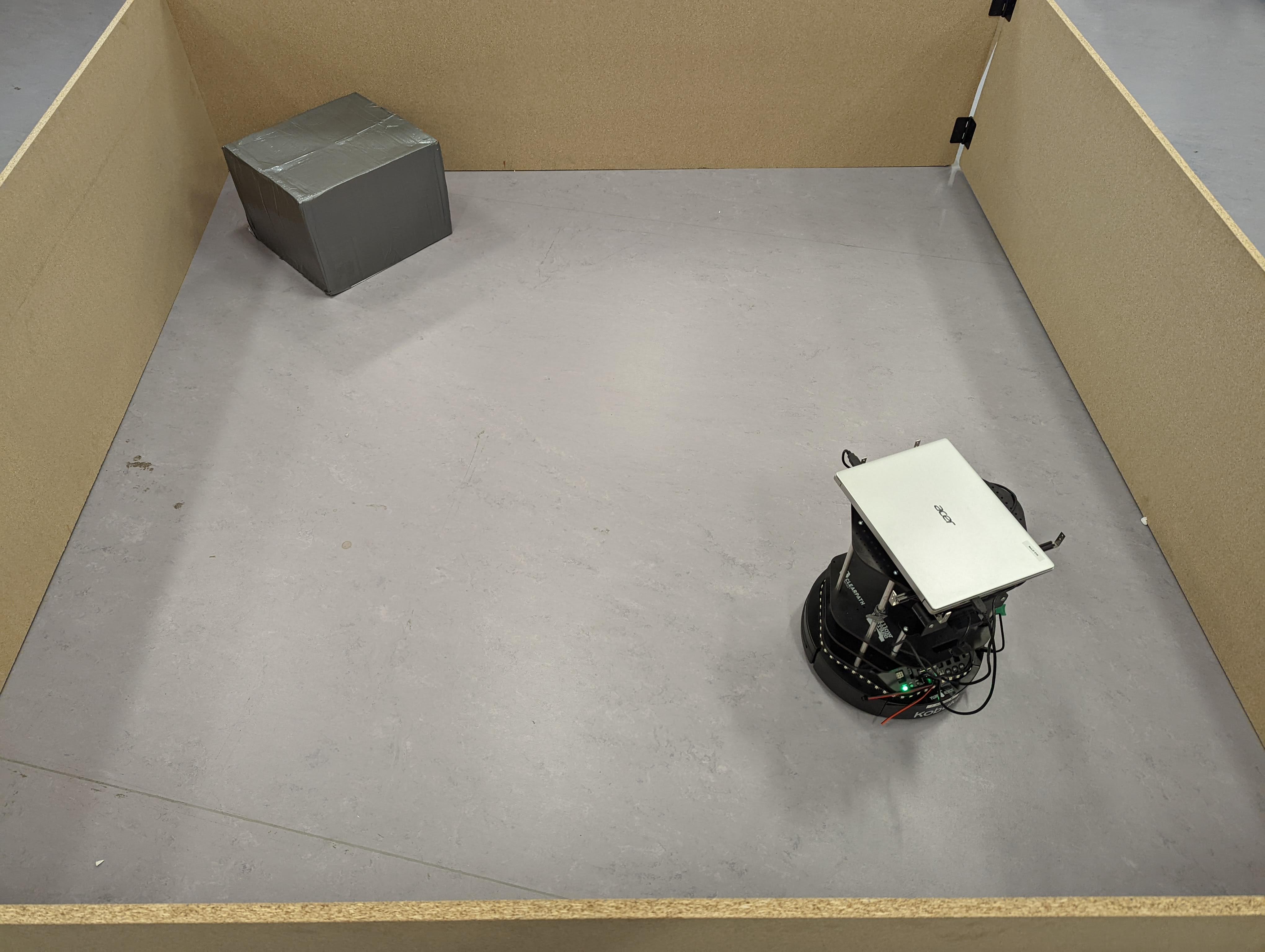}}
    \hspace{3mm}
    \subfloat[Observations in real-world \label{fig:real-obs}]
    {\includegraphics[width=0.23\textwidth, height=0.15\textwidth]{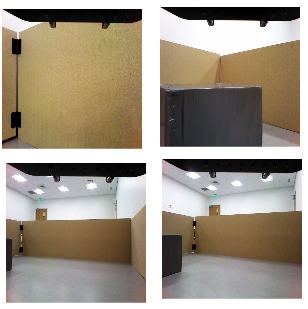}}
    \caption{ Real-world domain (left) and egocentric observations from the robot (right)}
    \label{fig:exp4_obs}
\end{figure}

To answer the fourth question, we evaluated our few-shot policy transfer approach on a Sim2Real task, where the goal of the robot in the real world was to reach the box as shown in Fig.~\ref{fig:real-world-domain}. The domain consisted of the robot navigating in an enclosed arena of dimensions $[2.45m, 2.45m]$. The dimensions of the goal object (box) are $[17in \times 19in \times 12in]$. We introduced a mismatch in the dimensions of the arena as well as the goal object to prevent an exact correspondence between the simulation and the real-world domains. For the real-world task, we made use of the TurtleBot~2 with an on-board laptop (running Ubuntu 18.04 and Robot Operating System Melodic~\cite{quigley2009ros}) which interfaces with a camera (Orbbec Astra). The laptop has an Intel(R) Core(TM) i7-1165G7 2.8GHz processor and 8GB RAM. The robot is equipped with an RPLIDAR A2M8 360$\degree$ Laser Scanner, but our observations consisted of only RGB images. Using the depth information will further enhance the observations, and can be used for tasks where depth information is necessary.

Along with a mismatch in the observations, we introduced a mismatch in the actions of the agents in the real-world and the simulated environment. In the real-world, the agent has access to four navigation actions, namely, move forward and backward ($0.04m$), and rotate clockwise and counter-clockwise ($\pi/100$). The dynamics of the source domain for this Sim2Real task are described in Section~\ref{sec:exp_1}. The static dataset for the real-world task consisted of $4000$ images taken using a random behavior policy on the robot. The entire dataset was collected in a span of 2 hours on the physical robot. Examples of real-world observations are shown in Fig.~\ref{fig:real-obs}\footnote{CycleGAN generated images in
Appendix~\href{https://github.com/shukla-yash/Few-Shot-Policy-Transfer/blob/main/appendix.pdf}{https://github.com/shukla-yash/Few-Shot-Policy-Transfer/blob/main/appendix.pdf}}. A subset of the images do not contain the goal object, and in order to reach the goal object, the robot needs to rotate to find out the location of the goal object.

Even with a significant mismatch in the observation and action spaces of the agent in the simulated domain and the real-world domain, we observed a $90\%$ accuracy of the robot in the real-world. We evaluated our few-shot policy transfer approach on $30$ episodes in the real-world, where the robot was able to reach the goal within the maximum number of permissible timesteps in $27$ out of the $30$ episodes. Our agent's latency was approximately 1 second (time taken for the agent to capture an image and generate an action after doing a forward pass).
In scenarios where the robot was unable to face the goal object, it learned to rotate itself to face the object successfully.
Once it was able to locate the goal object in its observation, it was able to navigate and reach the goal object.  However, in certain cases, the model was unable to differentiate between shadow caused by the walls and the box itself, because of the similarity in the RGB composition of the shadow and the box. In those scenarios, the robot either took longer to reach the goal or was unable to reach the goal in the predetermined number of timesteps. Here, enriching the observations with depth information will help the robot differentiate between the goal object and the shadow, and help in improving the accuracy. 

\section{Conclusion and Future Work} \label{sec:Conclusion}

In this work, we proposed a framework for few-shot policy transfer through observation mapping and behavior cloning. We used CycleGAN to generate the mapping between two dissimilar domains and then used behavior cloning to transfer the learned policy from one domain to another. We observed a successful few-shot transfer even when a limited number of interaction datapoints were available from the target domain. As the number of available datapoints from the target domain increases, the success rate performance increases. Additionally, when the task goal semantics and the action space parameters were changed in the target domain, we observed a $78\%$ few-shot success rate, which increased online fine-tuning in the target domain. Finally, we performed a Sim2Real transfer, by collecting a static dataset of images in the real-world domain, and then using our few-shot approach for policy transfer. We observed a goal-reaching success rate of $90\%$ in the real-world.  

Our framework for the few-shot policy transfer addressed a simple robot navigation scenario. For complicated tasks, generating a mapping between the observations in the two domains might require changes to the CycleGAN design. Additionally, for complicated tasks, we would like to use a curriculum-like setting, which would enable us to learn complex tasks incrementally, reducing the time required to learn in the real-world domain, and expanding to robotic manipulation settings.
In future work, we would like to address the problem of cross-domain knowledge transfer, where two domains do not have explicit mapping in their MDP representations but are semantically related. A potential approach for cross-domain knowledge transfer is to embed the task information in a latent world model and generate a mapping between the two domains. Furthermore, we would like to incorporate the history of the task's episode using an LSTM model.

\bibliographystyle{IEEEtran}
\balance
\bibliography{references}

\end{document}